\documentclass{article}

\PassOptionsToPackage{numbers, compress}{natbib}

\usepackage[preprint]{multimodal_em}

\usepackage[utf8]{inputenc} 
\usepackage[T1]{fontenc}    
\usepackage{url}            
\usepackage{booktabs}       
\usepackage{amsfonts}       
\usepackage{nicefrac}       
\usepackage{microtype}      
\usepackage{xcolor}         
\usepackage{graphicx}

\usepackage[margin=1em]{caption}

\definecolor{ccolor}{RGB}{140, 130, 247}
\usepackage[citecolor=ccolor, linkcolor=ccolor,colorlinks=true, urlcolor=ccolor]{hyperref}

\title{Brain encoding models based on multimodal transformers can transfer across language and vision}

\author{%
  Jerry Tang \\
  UT Austin \\
  \texttt{jerrytang@utexas.edu} \\
  \And
  Meng Du \\
  Intel Labs, UCLA \\
  \texttt{mengdu@ucla.edu} \\
  \And
  Vy A. Vo \\
  Intel Labs \\
  \texttt{vy.vo@intel.com} \\
  \And
  Vasudev Lal \\
  Intel Labs\\
  \texttt{vasudev.lal@intel.com} \\  
  \And
  Alexander G.~Huth \\
  UT Austin \\
  \texttt{huth@cs.utexas.edu}\\
}

\begin{document}

\maketitle

\begin{abstract}
Encoding models have been used to assess how the human brain represents concepts in language and vision. While language and vision rely on similar concept representations, current encoding models are typically trained and tested on brain responses to each modality in isolation. Recent advances in multimodal pretraining have produced transformers that can extract aligned representations of concepts in language and vision. In this work, we used representations from multimodal transformers to train encoding models that can transfer across fMRI responses to stories and movies. We found that encoding models trained on brain responses to one modality can successfully predict brain responses to the other modality, particularly in cortical regions that represent conceptual meaning. Further analysis of these encoding models revealed shared semantic dimensions that underlie concept representations in language and vision. Comparing encoding models trained using representations from multimodal and unimodal transformers, we found that multimodal transformers learn more aligned representations of concepts in language and vision. Our results demonstrate how multimodal transformers can provide insights into the brain’s capacity for multimodal processing.
\end{abstract}

\refstepcounter{section}
\label{introduction}

Encoding models predict brain responses from quantitative features of the stimuli that elicited them \cite{naselaris2011encoding}. In recent years, fitting encoding models to data from functional magnetic resonance imaging (fMRI) experiments has become a powerful approach for understanding information processing in the brain.  While encoding models are usually trained and tested on brain responses to a single stimulus modality, such as language \cite{Caucheteux2022BrainsAA, Goldstein2022SharedCP, jain2018incorporating, lebel2021voxelwise, Schrimpf2020TheNA, toneva2019interpreting, wehbe2014simultaneously} or vision \cite{eickenberg2017seeing, gucclu2015deep, huth2012continuous, naselaris2009bayesian, nishimoto2011reconstructing, kay2008identifying}, the human brain is remarkable in its ability to integrate information across multiple modalities. There is growing evidence that this capacity for multimodal processing is supported by aligned cortical representations of the same concepts in different modalities—for instance, hearing the sentence ``a dog chases a cat'' and seeing a dog chasing a cat may elicit similar patterns of brain activity \cite{binder2011neurobiology, devereux2013representational, fairhall2013brain, martin2016grapes, popham2021visual, tang2023semantic}.

\begin{figure}
  \centering
  \vspace{-1.5em}
  \includegraphics[width=\textwidth]{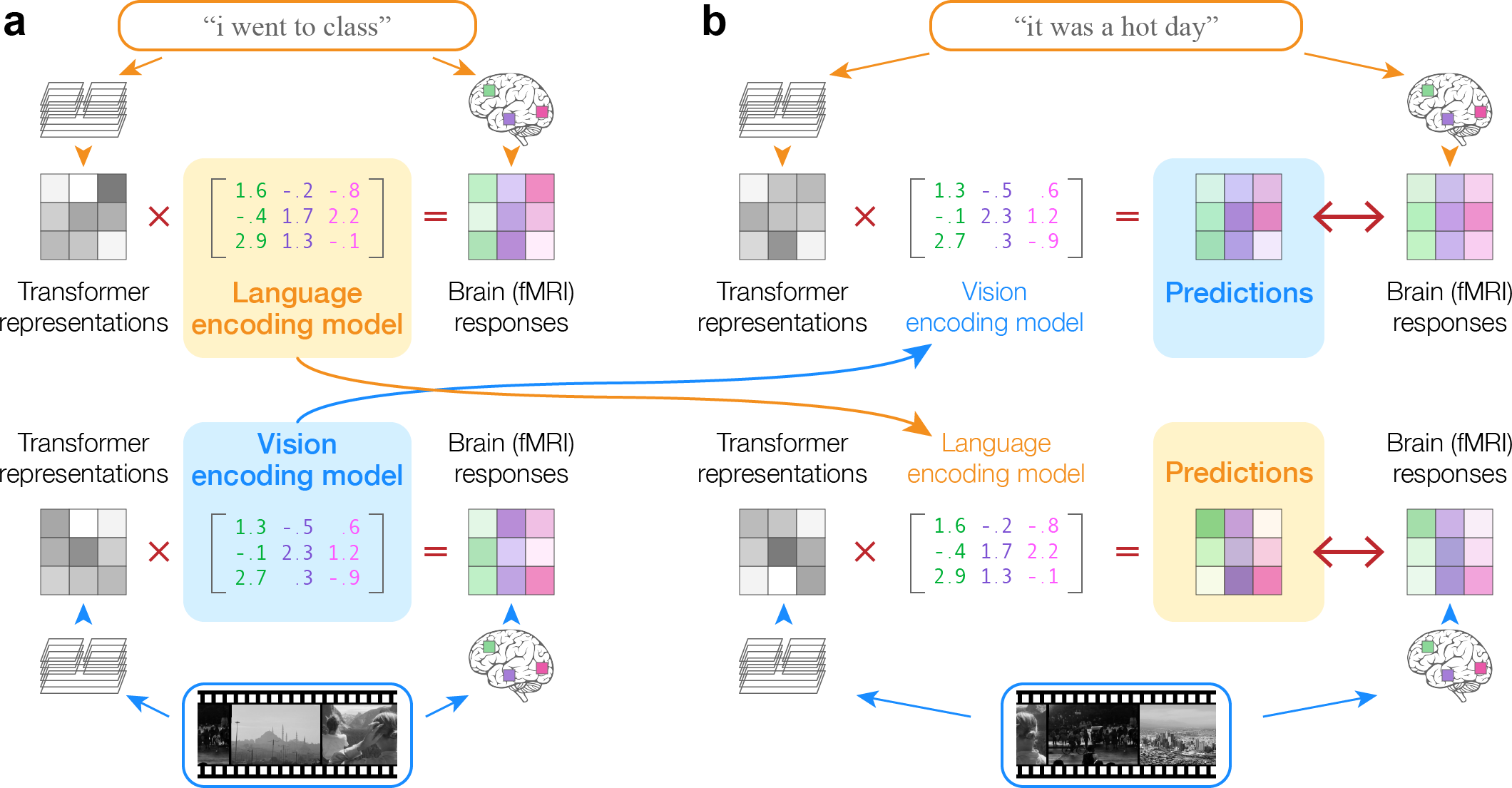}
  \caption{Cross-modality encoding model transfer. (\textbf{a}) Multimodal transformers were used to extract features of story and movie stimuli. Language encoding models were fit (using L2-regularized linear regression) to predict story fMRI responses from story stimuli; vision encoding models were fit to predict movie fMRI responses from movie stimuli. (\textbf{b}) As the two models share a representational space, language encoding models could be used to predict fMRI responses to movie stimuli, and vision encoding models to predict fMRI responses to story stimuli. Encoding model performance was quantified by the linear correlation between the predicted and the actual response time-courses in each voxel.}
  \vspace{-1.5em}
\label{fig:schematic}
\end{figure}

In this work, we investigated the alignment between language and visual representations in the brain by training encoding models on fMRI responses to one modality and testing them on fMRI responses to the other modality. Encoding models that successfully transfer across modalities can provide insights into how the two modalities are related \cite{popham2021visual}. Although previous work has compared language and vision encoding models, human annotations were required to map language and visual stimuli into a shared semantic space \cite{popham2021visual}. To our knowledge, cross-modality transfer has yet to be demonstrated using encoding models trained on stimulus-computable features that capture the rich connections between language and vision.

One way to extract aligned features of language and visual stimuli is using transformer models trained on multimodal objectives like image-text matching \cite{Chen2019UNITERUI, kim2021vilt, liu2021kd, lu2019vilbert, radford2021learning, tan2019lxmert, xu2022bridge}. Recent studies have shown that multimodal transformers outperform unimodal transformers at modeling brain responses to language and visual stimuli, suggesting that multimodal training enables models to learn more brain-like representations  \cite{lu2022multimodal, wang2022incorporating}. However, these studies do not assess whether representations from multimodal transformers can be used to train encoding models that transfer across modalities. Since multimodal transformers are trained to process paired language and visual inputs, the representations learned for a concept in language could be correlated with the representations learned for that concept in vision. This alignment between language and visual representations in multimodal transformers could facilitate the transfer of encoding models across modalities, which relies on the alignment between language and visual representations in the brain.

To test this, we used the BridgeTower \cite{xu2022bridge} multimodal transformer to model fMRI responses to naturalistic stories \cite{huth2016natural} and movies \cite{huth2012continuous}. We separately obtained quantitative features of story and movie stimuli by extracting latent representations from BridgeTower. We estimated language encoding models using story features and story-fMRI responses, and vision encoding models using movie features and movie-fMRI responses (Figure \ref{fig:schematic}a). We evaluated how well the language encoding models can predict movie-fMRI responses from movie features ($story \rightarrow movie$) and how well the vision encoding models can predict story-fMRI responses from story features ($movie \rightarrow story$) (Figure \ref{fig:schematic}b). We compared this to how well the language encoding models can predict story-fMRI responses from story features ($story \rightarrow story$) and how well the vision encoding models can predict movie-fMRI responses from movie features ($movie \rightarrow movie$).

We found that encoding models trained on brain responses to one modality could accurately predict brain responses to the other modality. In many brain regions outside of sensory and motor cortex, $story \rightarrow movie$ performance approached $movie \rightarrow movie$ performance, suggesting that these regions encode highly similar representations of concepts in language and vision. To assess these representations, we performed principal components analysis on the encoding model weights and identified semantic dimensions that are shared between concept representations in language and vision. Finally, we found that cross-modality performance was higher for features extracted from multimodal transformers than for linearly aligned features extracted from unimodal transformers. Our results characterize how concepts in language and vision are aligned in the brain and demonstrate that multimodal transformers can learn representations that reflect this alignment.

\section{Multimodal transformers}
\label{transformers}

Multimodal transformers are trained on paired language and visual data to perform self-supervised tasks such as image-text matching. Typically, these models are used to extract representations of paired language and visual input for downstream tasks such as visual question answering. However, since multimodal training objectives may impose some degree of alignment between language and visual tokens for the same concept, these models could also be used to extract aligned representations of language and visual input in isolation. For instance, latent representations extracted from a model given the sentence “a dog chases a cat” may be correlated with latent representations extracted from the model given a picture of a dog chasing a cat.

\subsection{BridgeTower}

In this study, we extracted stimulus features using a pretrained BridgeTower model \cite{xu2022bridge}. BridgeTower is a vision-language transformer trained on image-caption pairs from the Conceptual Captions \cite{sharma-etal-2018-conceptual}, SBU Captions \cite{NIPS2011_5dd9db5e}, MS COCO Captions \cite{Lin2014MicrosoftCC}, and Visual Genome \cite{Krishna2016VisualGC} datasets. For each image-caption pair, the caption is processed using a language encoder initialized with pretrained RoBERTa parameters \cite{Liu2019RoBERTaAR} while the image is processed using a vision encoder initialized with pretrained ViT parameters \cite{Dosovitskiy2020AnII}. The early layers of BridgeTower process language and vision tokens independently, while the later layers of BridgeTower are cross-modal layers that process language and vision tokens together. Results are shown for the BridgeTower-Base model; corresponding results for the BridgeTower-Large model are shown in Appendix \ref{supp:btlarge}.

\label{featureextraction}

We used BridgeTower to extract features from the story (Section \ref{story_fmri}) and movie (Section \ref{movie_fmri}) stimuli that were used in the fMRI experiments. Each story and movie stimulus was separately processed using BridgeTower by running forward passes with input from one of the two modalities. Hidden representations were extracted from each layer of BridgeTower as it processed the inputs.

For stories, segments of transcripts were used as model inputs with no accompanying image inputs. A feature vector was obtained for every word by padding the target word with a context of 20 words both before and after. For movies, single frames were used as model inputs with no accompanying text inputs. Movies were presented at 15 frames per second, and a feature vector was obtained for every 2-second segment by averaging latent representations of every 30 frames. This was done on a node with 10 Intel Xeon Platinum 8180 CPUs and an Nvidia Quadro RTX 6000 GPU.

\subsection{Alignment of feature spaces}
\label{featurealignment}

Transformers compute representations for each layer by attending to different combinations of the input tokens. While multimodal training tasks may require models to align language and visual tokens for the same concept, the nature of this alignment depends on the type of attention mechanism used to combine language and visual tokens \cite{hendricks2021decoupling}. BridgeTower uses a co-attention mechanism wherein language and visual tokens are passed through different projection matrices, and query vectors from each modality are only scored against key vectors from the other modality. As a consequence, the language and visual feature spaces extracted from each layer of BridgeTower may differ up to a linear transformation.

To correct for these potential transformations, we used the Flickr30K dataset \cite{young-etal-2014-image}—which consists of paired captions and images—to estimate linear transformation matrices that explicitly align the BridgeTower feature spaces. We used BridgeTower to separately extract language features of each caption and visual features of each image. We then used L2-regularized linear regression to estimate $image \rightarrow caption$ matrices that predict each language feature from the visual features, and $caption \rightarrow image$ matrices that predict each visual feature from the language features. Before using the language encoding model to predict fMRI responses to movies, we first used the $image \rightarrow caption$ matrix to project the movie features into the language feature space. Similarly, before using the vision encoding model to predict fMRI responses to stories, we first used the $caption \rightarrow image$ matrix to project the story features into the visual feature space.

\section{fMRI experiments}
\label{fmri}

We analyzed publicly available fMRI data from five subjects (2 female, 3 male) who participated in a story listening experiment and a movie watching experiment \cite{popham2021visual}. Blood-oxygen level dependent (BOLD) brain signals were recorded using gradient-echo EPI on a a 3T Siemens TIM Trio scanner at the UC Berkeley Brain Imaging Center with a 32-channel volume coil, TR = 2.0045 seconds, TE = 31 ms, flip angle = 70 degrees, voxel size = 2.24 $\times$ 2.24 $\times$ 4.1 mm (slice thickness = 3.5 mm with 18 percent slice gap), matrix size = 100 $\times$ 100, and 30 axial slices. All experiments and subject compensation were approved by the UC Berkeley Committee for the Protection of Human Subjects.

\subsection{Story experiment}
\label{story_fmri}

Stimuli for the story experiment consisted of 10 naturally spoken narrative stories from The Moth Radio Hour ranging from 10 to 15 minutes and totaling just over 2 hours \cite{huth2016natural}. The stories were presented over Sensimetrics S14 headphones. Subjects were instructed to listen to the stories with their eyes closed. Each story was played during a single fMRI scan.

\subsection{Movie experiment}
\label{movie_fmri}

Stimuli for the movie experiment consisted of 12 videos totaling 2 hours \cite{huth2012continuous, nishimoto2011reconstructing} Each video was made by concatenating a sequence of 10-20 s clips from movies drawn from the \href{https://trailers.apple.com/}{Apple QuickTime HD gallery} and \href{https://www.youtube.com/}{YouTube}. The videos were presented at 15 frames per second. Subjects were instructed to fixate on a dot at the center of the screen. Each video was played during a single fMRI scan.

\section{Voxelwise encoding models}
\label{encoding}

Voxelwise fMRI encoding models learn a mapping from stimuli to the brain responses that they elicit in each individual subject \cite{huth2016natural}. Brain images recorded at times $t = 1...T$ are given by $y(t) \in \mathbb{R}^m$ where $m$ is the number of voxels in the cerebral cortex. Responses for one subject are represented by the response matrices $Y_{story} \in \mathbb{R}^{T_{story} \times m}$ and $Y_{movie} \in \mathbb{R}^{T_{movie} \times m}$.

The story and movie features were resampled to the fMRI acquisition times using a Lanczos filter. To account for the hemodynamic response, a finite impulse response model with 4 delays (2, 4, 6, and 8 seconds) was applied to the downsampled features. This resulted in the delayed stimulus matrices $X_{story} \in \mathbb{R}^{T_{story} \times 4k}$ and $X_{movie} \in \mathbb{R}^{T_{movie} \times 4k}$ where $k = 768$ is the dimensionality of the BridgeTower features.

We modeled the mapping between stimulus features and brain responses with a linear model $Y = X\beta$. Each column of $\beta$ represents the linear weights on the $4k$ delayed features for each voxel. The weights $\beta$ were estimated using L2-regularized linear regression. Regularization parameters were independently selected for each voxel using 50 iterations of a cross-validation procedure. This was done on a workstation with an Intel Core i9-7900X CPU.

\subsection{Evaluation}

Encoding models were evaluated by predicting the response matrices $Y_{test}$ from the stimulus matrices $X_{test}$ for stimuli that were excluded from model estimation. Prediction performance for each voxel is quantified by the linear correlation between the predicted and actual response time-courses.

To quantify $source \rightarrow target$ performance from a source modality to a target modality, we estimated encoding models using all source scans and evaluated prediction performance on each target scan. We averaged linear correlations across the target scans to obtain a score $r_{source \rightarrow target}$ for each voxel.

We compared this cross-modality performance against the within-modality performance of an encoding model trained on the target modality. To quantify $target \rightarrow target$ performance, we held out each target scan, estimated encoding models using the remaining target scans, and evaluated prediction performance on the held out target scan. We averaged linear correlations across the held out target scans to obtain a score $r_{target \rightarrow target}$ for each voxel.

We separately identified voxels with statistically significant $story \rightarrow story$ and $movie \rightarrow movie$ performance using a blockwise permutation test described in Appendix \ref{supp:significance}.

\subsection{Layer selection}

Separate encoding models were trained using stimulus features extracted from each layer of BridgeTower. We summarized performance across layers by estimating the best layer for each voxel using a bootstrap procedure. For each test scan, we estimated the best layer for each voxel based on mean prediction performance across the remaining test scans. We then used the selected layer for each voxel to compute prediction performance for that voxel on the held out test scan. We used this procedure for all analyses unless noted otherwise.

\section{Results}
\label{results}

We separately estimated language and vision encoding models for each subject. We used these models to compute $r_{story \rightarrow movie}$, $r_{movie \rightarrow story}$, $r_{story \rightarrow story}$, and $r_{movie \rightarrow movie}$ scores for each voxel.

\subsection{Cross-modality performance}

Cross-modality performance was visualized by projecting $r_{story \rightarrow movie}$ and $r_{movie \rightarrow story}$ scores for each voxel in one subject onto a flattened cortical surface (Figure \ref{fig:crossmodality}a; see Appendix \ref{supp:flatmaps} for flatmaps for other subjects). We found positive $r_{story \rightarrow movie}$ and $r_{movie \rightarrow story}$ scores in many parietal, temporal, and frontal regions, which have previously been shown to represent the meaning of concepts in language \cite{huth2016natural} and vision \cite{huth2012continuous}. The high $story \rightarrow movie$ performance and positive (albeit lower) $movie \rightarrow story$ performance suggest that these voxels have similar tuning for the same concepts across modalities \cite{popham2021visual}.

\begin{figure}
  \centering
  \vspace{-1.5em}
  \includegraphics[width=\textwidth]{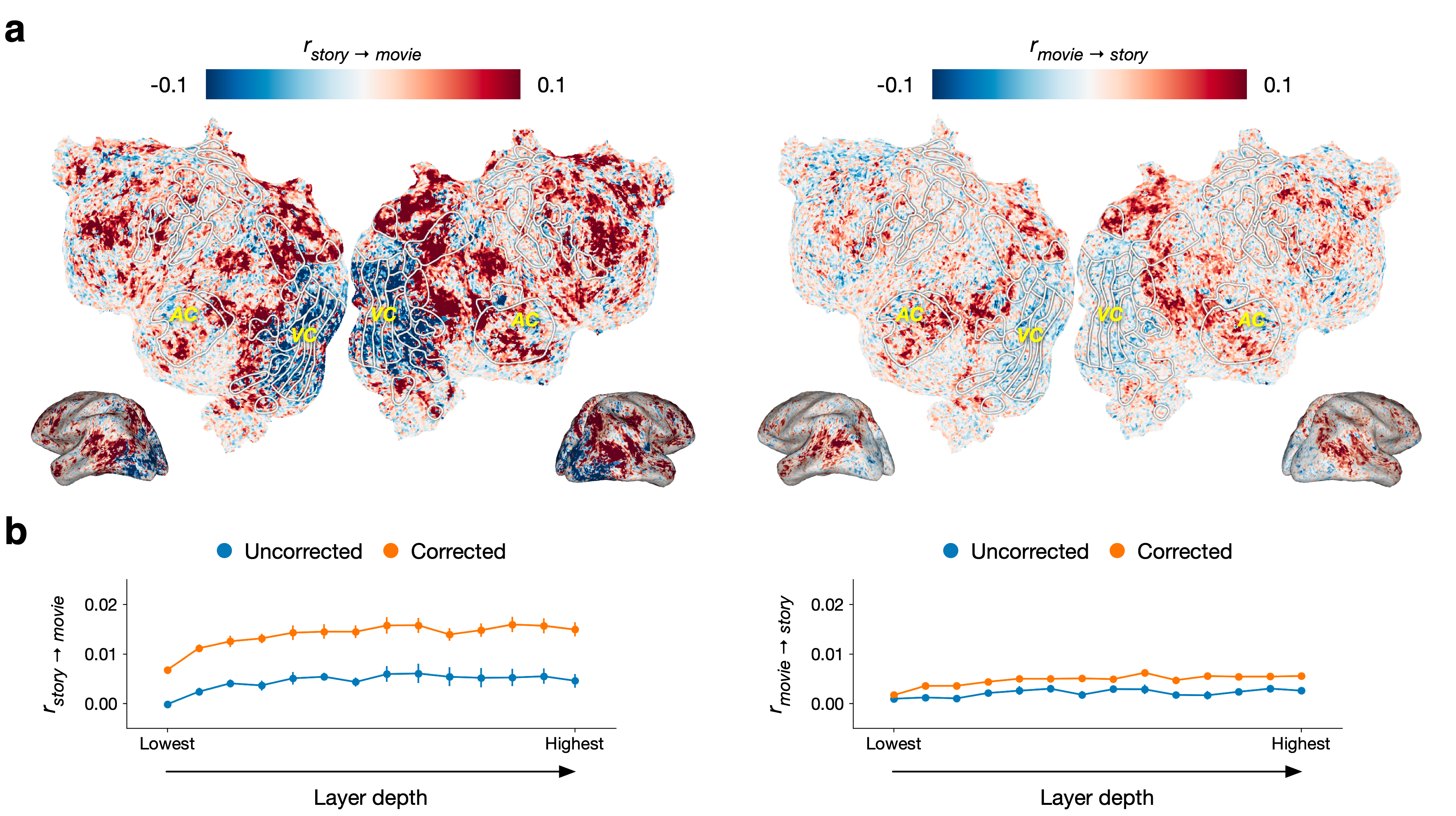}
  \caption{Cross-modality prediction performance. Encoding models estimated on brain responses to one modality were evaluated on brain responses to the other modality. Cross-modality performance is measured by the linear correlation ($r$) between predicted and actual responses. (\textbf{a}) $r_{story \rightarrow movie}$ and $r_{movie \rightarrow story}$ scores for each voxel in one subject are displayed on that subject's cortical surface. Voxels appear red if performance is positive, blue if performance is negative, and white if performance is zero. White outlines show regions of interest (ROIs) identified using separate localizer data. VC and AC denote visual cortex and auditory cortex. $r_{story \rightarrow movie}$ and $r_{movie \rightarrow story}$ scores were positive in regions that have previously been found to represent the meaning of concepts in language and vision, but negative in visual cortex. (\textbf{b}) Prediction performance for each layer of the model. Negative correlations were corrected by using held-out data to fit a one-parameter model for each voxel that predicts whether the encoding model weights should be negated before computing transfer performance. Correlations were averaged across voxels and then across subjects. Error bars indicate standard error of the mean across subjects.}
  \vspace{-1.5em}
\label{fig:crossmodality}
\end{figure}

Conversely, we found negative $r_{story \rightarrow movie}$ and $r_{movie \rightarrow story}$ scores in visual cortex. Previous studies have reported that the tuning to perceptual information in visual cortex may be inverted during conceptual processing in the absence of perception \cite{buchsbaum2012neural, szinte2020visual}. If there is systematically inverted tuning between language and vision, it should be possible to first estimate which voxels would have negative cross-modality performance using separate validation data, and then multiply their weights by $-1$ before computing $r_{story \rightarrow movie}$ and $r_{movie \rightarrow story}$ on test data. We performed this correction using a bootstrap procedure across the test scans. For each test scan, we estimated voxels with inverted tuning based on the mean prediction performance across the remaining test scans. We then multiplied the weights of these voxels by $-1$ before computing prediction performance on the held out test scan.

Figure \ref{fig:crossmodality}b shows $story \rightarrow movie$ and $movie \rightarrow story$ performance across cortex before and after this correction. We summarized the performance for each layer of BridgeTower by averaging the linear correlations across all cortical voxels and subjects. Across layers, the correction significantly improved $story \rightarrow movie$ performance (one-sided paired t-test; $p < 0.05$, $t(4) = 7.5295$, $\overline{r}_{corrected} = 0.0230$, $\overline{r}_{uncorrected} = 0.0053$) and $movie \rightarrow story$ performance (one-sided paired t-test; $p < 0.05$, $t(4) = 6.6356$, $\overline{r}_{corrected} = 0.0097$, $\overline{r}_{uncorrected} = 0.0031$), providing evidence for systematically inverted tuning for the same concepts across modalities.

\subsection{Comparing cross-modality and within-modality performance}

\begin{figure}
  \centering
  \vspace{-1.5em}
  \includegraphics[width=\textwidth]{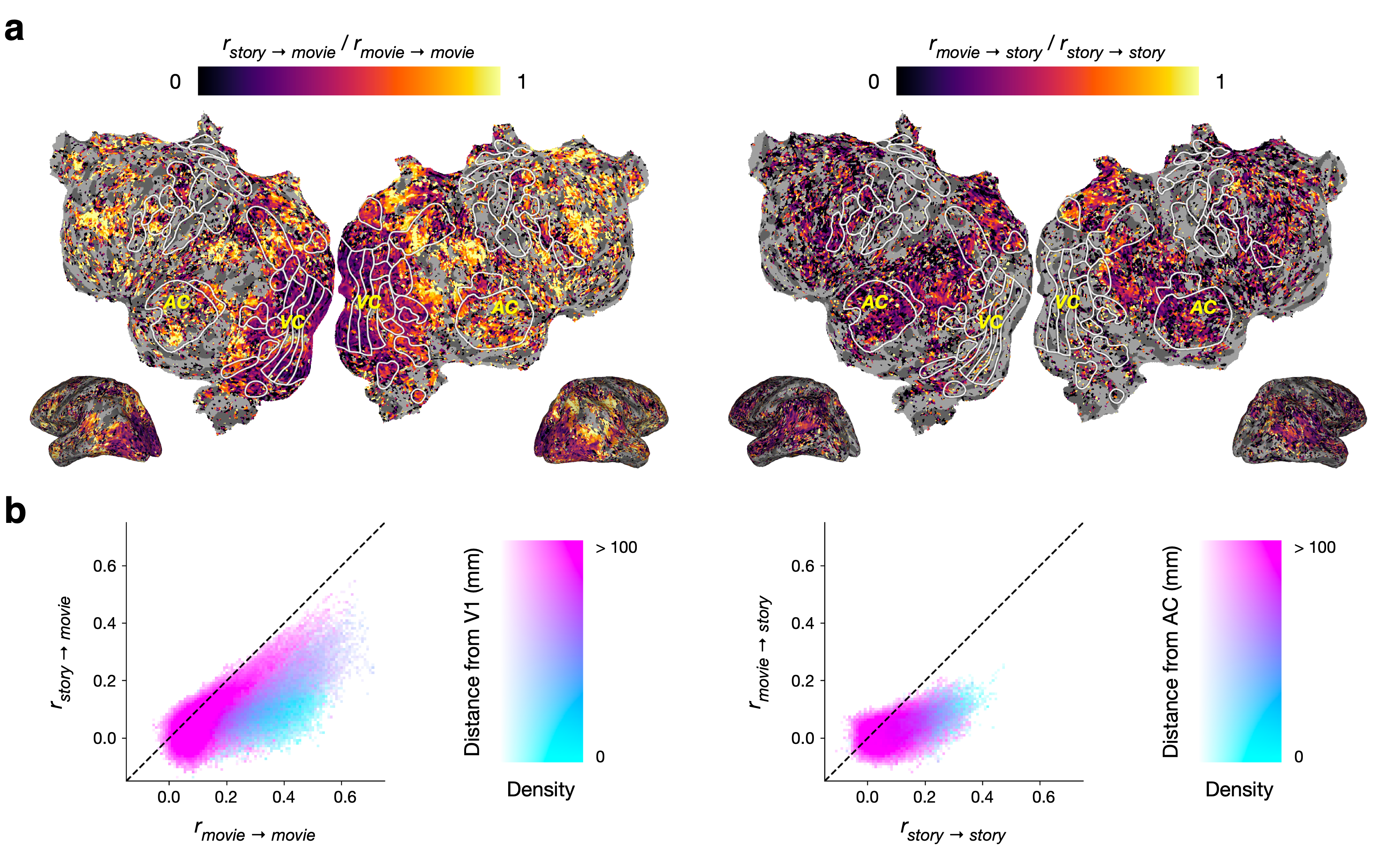}
  \caption{Comparing cross- and within-modality prediction performance. Cross-modality scores were compared against within-modality scores in voxels with statistically significant within-modality scores. (\textbf{a}) Cross- and within-modality scores for each voxel in one subject are projected onto the subject's flattened cortical surface. Voxels appear dark if cross-modality performance is much lower than within-modality performance, and bright if cross-modality performance approaches within-modality performance. Only well-predicted voxels under the within-modality model (q(FDR) < 0.05, one-sided permutation test) are shown. (\textbf{b}) Histograms compare cross-modality performance to within-modality performance. For responses to movies, $r_{story \rightarrow movie}$ scores were much lower than $r_{movie \rightarrow movie}$ near visual cortex but approached $r_{movie \rightarrow movie}$ scores in other regions. For responses to stories, $r_{movie \rightarrow story}$ scores were generally much lower than $r_{story \rightarrow story}$ scores across cortex.}
  \vspace{-1.5em}
\label{fig:withinmodality}
\end{figure}

While the previous analysis identified voxels with similar tuning for concepts in language and vision, it did not characterize the extent of this cross-modal similarity. To do this, we next compared cross-modality performance to within-modality performance for each voxel (Figure \ref{fig:withinmodality}a).

To quantify the amount of information that the language encoding model learns about tuning for movies, we divided $r_{story \rightarrow movie}$ by $r_{movie \rightarrow movie}$ for each voxel. If the movie responses in a voxel are well-predicted by the vision encoding model but poorly predicted by the language encoding model, this value should be low. Conversely, if the movie responses are predicted about as well using both the vision and language encoding models, then this value should be close to $1$. In visual cortex, which represents structural features of visual stimuli \cite{nishimoto2011reconstructing}, the language encoding model performed much worse than the vision encoding model at predicting responses to movies. In significantly predicted voxels outside of visual cortex, which represent the meaning of visual stimuli \cite{nunez2019voxelwise, huth2012continuous, popham2021visual}, the language encoding model often approached the performance of the vision encoding model (Figure \ref{fig:withinmodality}b).

Similarly, to quantify the amount of information that the vision encoding model learns about tuning for stories, we divided $r_{movie \rightarrow story}$ by $r_{story \rightarrow story}$ for each voxel. In auditory cortex, which represents acoustic and articulatory features of language stimuli \cite{de2017hierarchical}, the vision encoding model performed much worse than the language encoding model at predicting brain responses to stories. In some significantly predicted voxels outside of auditory cortex, which have been shown to represent the meaning of language stimuli \cite{de2017hierarchical, huth2016natural}, the vision encoding model performed relatively better, but still did not approach the performance of the language encoding model (Figure \ref{fig:withinmodality}b).

These results suggest that visual tuning can often be estimated solely based on how a voxel responds to stories, while it is much harder to estimate language tuning solely based on how a voxel responds to movies. One potential confound that could contribute to this asymmetry is that the story stimuli contain both concrete concepts (such as places) and abstract concepts (such as emotions) while the movie stimuli mostly contain concrete concepts. Another potential confound is that the story stimuli contain information at a longer timescale than the movie stimuli, which consist only of 10-20 second clips. To isolate whether the asymmetry in Figure \ref{fig:withinmodality} is driven by differences between language and visual representations in the brain, future work could use story and movie stimuli that are matched in terms of semantics and timescale coverage.

\subsection{Encoding model principal components}

The earlier analyses showed that encoding models can readily transfer across modalities, at least in the direction from language to vision. But what kind of information is it that these cross-modal models are capturing? To understand the semantic dimensions that underlie the shared tuning for concepts in language and vision, we next examined the principal components of the encoding model weights. We applied principal components analysis to the language encoding model weights of the top 10,000 voxels for each subject \cite{huth2016natural}, which produced 768 orthogonal principal components (PCs) that are ordered by the amount of variance they explain across the voxels. We projected stimulus features onto each PC to interpret the semantic dimension that the PC captures, and we projected encoding model weights onto each PC to assess how the corresponding semantic dimension is represented across cortex. We estimated encoding models using layer 8 of BridgeTower, which has the highest average performance across cortex, and did not correct for negative cross-modality scores.

\begin{figure}
  \centering
  \vspace{-1.5em}
  \includegraphics[width=\textwidth]{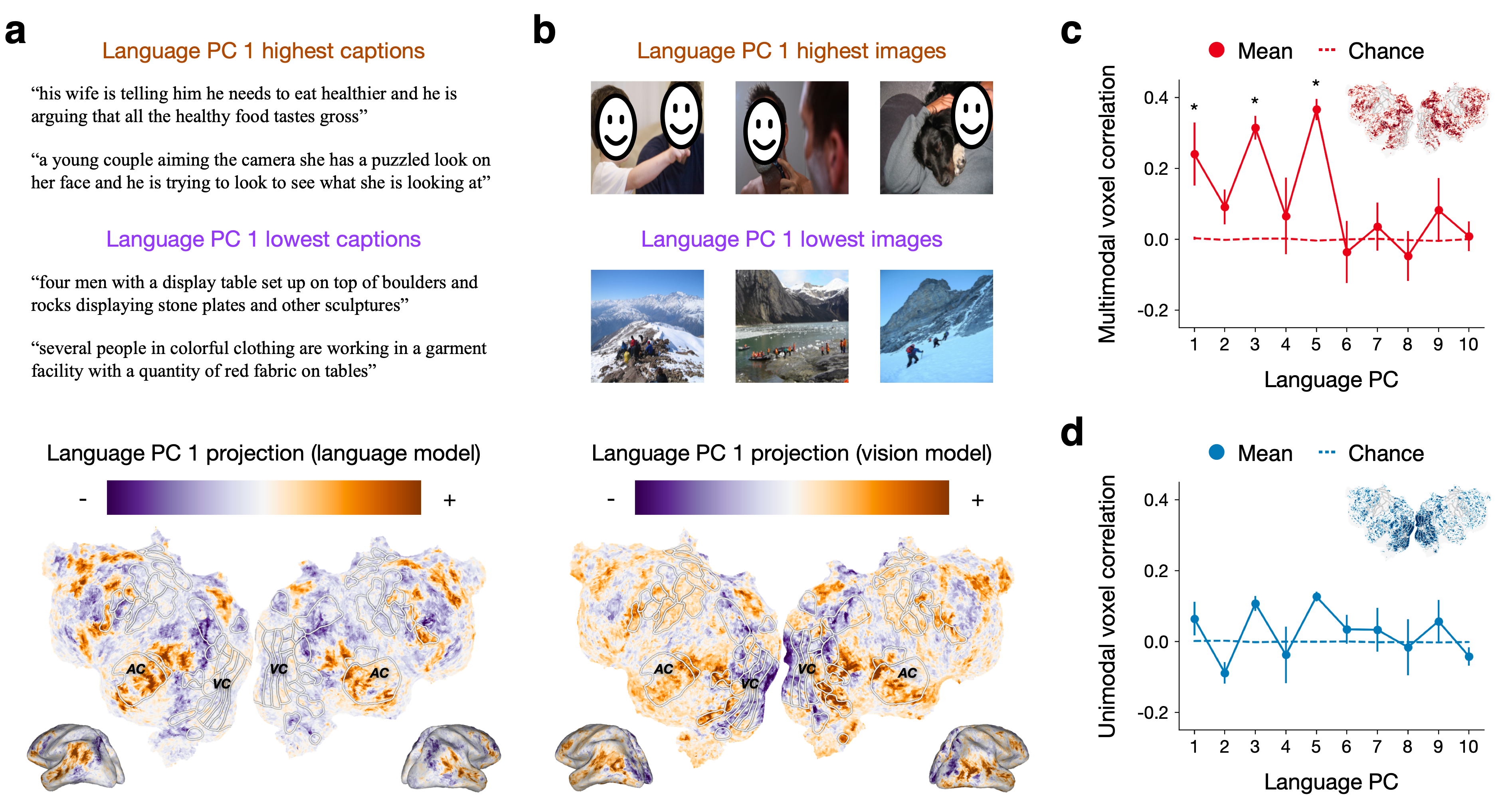}
  \caption{Encoding model principal components. Principal components analysis identified the first 10 principal components (PCs) of language encoding model weights. (\textbf{a}) Each caption in Flickr30k was projected onto language PC 1. This PC distinguishes captions that refer to people and social interactions—which are represented in inferior parietal cortex, precuneus, temporal cortex, and frontal cortex—from places and objects—which are represented in superior parietal cortex and middle frontal cortex. (\textbf{b}) Each image in Flickr30k was projected onto language PC 1. Here language PC 1  distinguishes images of people—which are represented in EBA, OFA, FFA, inferior parietal cortex, precuneus, temporal cortex, and frontal cortex—from images of places—which are represented in superior parietal cortex and middle frontal cortex. (\textbf{c}) In voxels that were well predicted by both the language and the vision encoding models (red on inset flatmap), projections of language and vision encoding model weights were significantly correlated (*) for several language PCs, indicating semantic dimensions that are shared between language and visual representations. (\textbf{d}) In voxels that were well predicted by only the language or the vision encoding models (blue on the inset flatmap), projections of language and vision encoding model weights were not significantly correlated for any language PCs.}
  \vspace{-1.5em}
\label{fig:pca}
\end{figure}

Projecting Flickr30k caption features onto the first PC of the language encoding model weights (language PC 1), we found that phrases with positive language PC 1 projections tend to refer to people and social interactions, while phrases with negative language PC 1 projections tend to refer to places and objects (Figure \ref{fig:pca}a). Projecting the language encoding model weights onto language PC 1, we found that voxels with positive projections were mostly located in inferior parietal cortex, precuneus, temporal cortex, and regions of frontal cortex; voxels with negative projections were mostly located in superior parietal cortex and middle frontal cortex. These findings are consistent with previous studies that mapped how different concepts in language are represented across cortex \cite{huth2016natural, deniz2019representation}.

Since our previous results show that many voxels have shared tuning for concepts in language and vision, the semantic dimensions captured by the language PCs may also underlie the space of visual representations. Projecting Flickr30k image features onto language PC 1, we found that images with positive language PC 1 projections tend to contain people, while images with negative language PC 1 projections tend to contain places such as natural scenes (Figure \ref{fig:pca}b). Projecting the vision encoding model weights onto language PC 1, we found similar patterns to the language encoding model projections outside of visual cortex. However, we additionally found voxels with positive projections in visual cortex regions known to represent faces (OFA, FFA) and body parts (EBA) in vision. These results suggest that the semantic dimension captured by language PC 1 is partially shared between language and visual representations.

We next quantified the degree to which each of the top 10 language PCs is shared between language and visual representations. For each PC, we spatially correlated the projections of the language and the vision encoding model weights. We separately computed spatial correlations across multimodal voxels that were well predicted by both the language and vision encoding models—operationalized as the 10,000 voxels with the highest $\min(r_{story \rightarrow story}, r_{movie \rightarrow movie})$—as well as across unimodal voxels that were well predicted by either the language or vision encoding model but not both—operationalized as the 10,000 remaining voxels with the highest $\max(r_{story \rightarrow story}, r_{movie \rightarrow movie})$. The spatial correlations quantify how similarly the voxels represent the semantic dimension captured by each PC.

We tested the significance of these correlations using a blockwise permutation test (Appendix \ref{supp:significance}). For multimodal voxels (Figure \ref{fig:pca}c), the projections of the language and the vision encoding model weights were significantly correlated for language PCs 1, 3, and 5 (q(FDR) < 0.05; see Appendix \ref{supp:pc} for further analyses). For unimodal voxels (Figure \ref{fig:pca}d), the projections of the language and the vision encoding model weights were not significantly correlated for any of the language PCs.

\subsection{Comparing transfer performance using multimodal and unimodal transformers}

Finally, we isolated the effects of multimodal training on cross-modality performance. To provide a unimodal baseline, we estimated language encoding models using RoBERTa \cite{Liu2019RoBERTaAR} and vision encoding models using ViT \cite{Dosovitskiy2020AnII}. Since these unimodal transformers were used to initialize BridgeTower, they provide a baseline for how well language and visual features are aligned prior to multimodal training.

\begin{figure}
  \centering
  \vspace{-1.5em}
  \includegraphics[width=\textwidth]{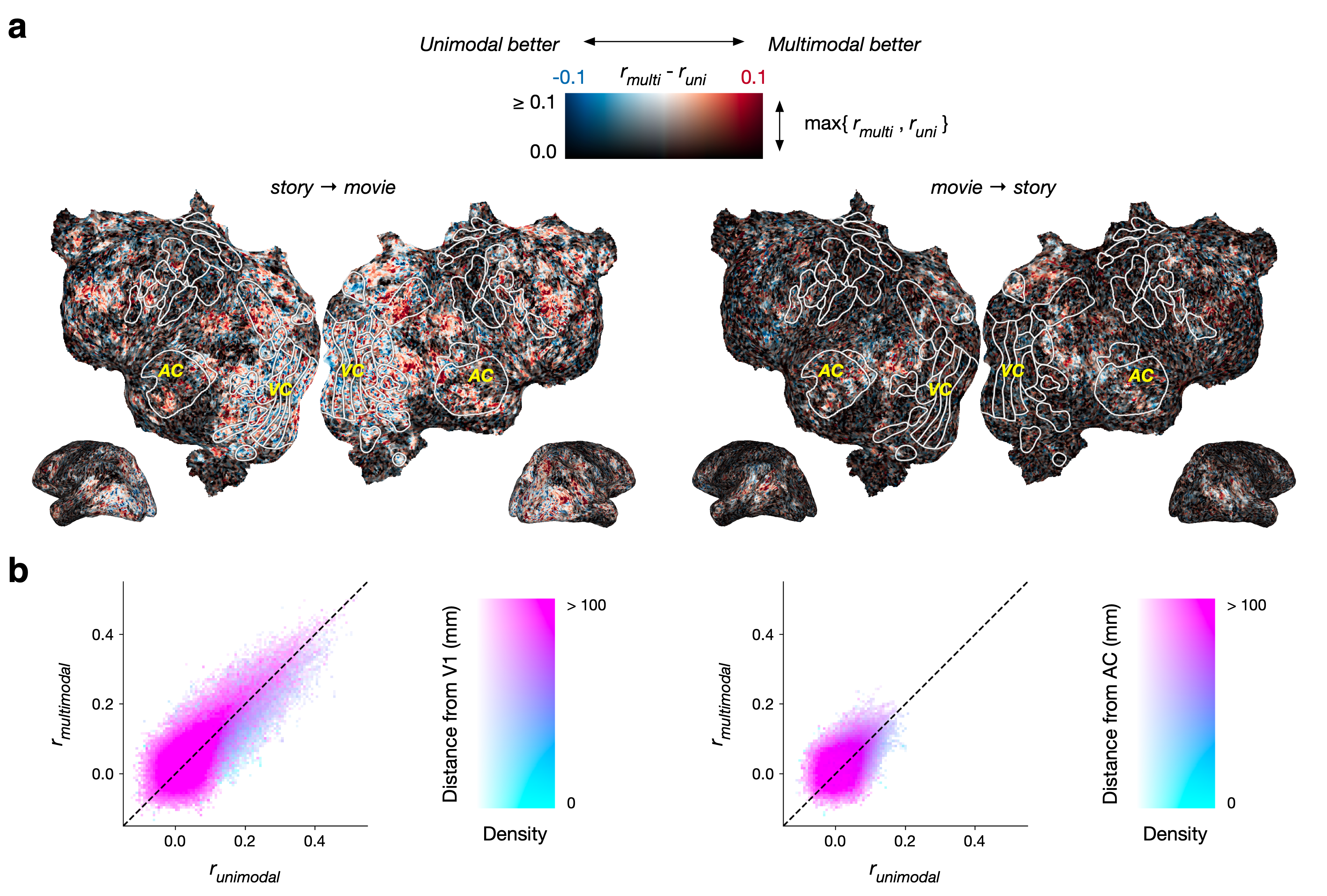}
  \caption{Transfer performance using features from multimodal and unimodal transformers. Cross-modality performance was compared between multimodal encoding models that extract features using BridgeTower and unimodal encoding models that extract features using RoBERTa and ViT. (\textbf{a}) The difference between multimodal performance and unimodal performance for each voxel in one subject is projected on the subject's flattened cortical surface. Voxels appear red if they are better predicted by multimodal features, blue if they are better predicted by unimodal features, white if they are well predicted by both, and black for neither. (\textbf{b}) Histograms compare multimodal performance to unimodal performance. For $story \rightarrow movie$ transfer, multimodal features outperform unimodal features in regions outside of visual cortex.}
  \vspace{-1.5em}
\label{fig:multivsuni}
\end{figure}

To perform cross-modal transfer with features from the unimodal transformers, we first estimated linear alignment matrices (Section \ref{featurealignment}) on the Flickr30k dataset. We estimated $image \rightarrow caption$ matrices that predict each RoBERTa language feature from the ViT visual features, and $caption \rightarrow image$ matrices that predict each ViT visual feature from the RoBERTa language features. We then used these alignment matrices to evaluate how well a RoBERTa language encoding model can predict movie-fMRI responses using ViT movie features, and how well a ViT vision encoding model can predict story-fMRI responses using RoBERTa story features. 

Across cortex, we found that multimodal features led to significantly higher $story \rightarrow movie$ performance (one-sided paired t-test; $p < 0.05$, $t(4) = 2.1377$, $\overline{r}_{multimodal} = 0.0230$, $\overline{r}_{unimodal} = 0.0219$) and $movie \rightarrow story$ performance (one-sided paired t-test; $p < 0.05$, $t(4) = 5.3746$, $\overline{r}_{multimodal} = 0.0097$, $\overline{r}_{unimodal} = 0.0074$) than the unimodal features (Figure \ref{fig:multivsuni}a). Specifically, multimodal features led to higher $story \rightarrow movie$ performance outside of visual cortex, and higher $movie \rightarrow story$ performance outside of auditory cortex (Figure \ref{fig:multivsuni}b). These results suggest that the multimodal training objectives induce the BridgeTower model to learn more complex connections between language and visual representations than a simple linear alignment between modalities.

\section{Discussion}

Our study demonstrates that encoding models trained on brain responses to language or visual stimuli can be used to predict brain responses to stimuli in the other modality, indicating similar conceptual representations of language and visual stimuli in the brain \cite{tang2023semantic, popham2021visual}. Our analyses identified the regions in which these representations are aligned, as well as the semantic dimensions underlying this alignment. Notably, however, while tuning for concepts in language and vision is positively correlated in most regions outside of visual cortex, it is \textit{negatively} correlated in visual cortex. Understanding the nature of this inverted tuning is an important direction for future work that could provide deeper insights into the relationship between language and vision \cite{buchsbaum2012neural, szinte2020visual}. 

To estimate the cross-modal encoding models, we used the BridgeTower multimodal transformer to extract features of the story and movie stimuli. The successful transfer performance demonstrates that multimodal transformers learn aligned representations of language and visual input. Moreover, stimulus features extracted from multimodal transformers led to better cross-modality performance than linearly aligned stimulus features extracted from unimodal transformers, suggesting that the multimodal training tasks enable BridgeTower to learn connections between language and visual concepts that go beyond a simple linear alignment between unimodal representations.

\begin{ack}
 This work was supported by the National Institute on Deafness and Other Communication Disorders under award number 1R01DC020088-001 (A.G.H.), the Whitehall Foundation (A.G.H.), the Alfred P. Sloan Foundation (A.G.H.) and the Burroughs Wellcome Fund (A.G.H.).
\end{ack}

\bibliographystyle{unsrt}
\bibliography{multimodal_em}


\newpage

\appendix

\section{Significance testing}
\label{supp:significance}

\subsection{Voxel performance significance test}

We computed $story \rightarrow story$ and $movie \rightarrow movie$ scores for each voxel by taking the linear correlation between the predicted response time course and the actual response time course. We separately identified voxels with statistically significant $story \rightarrow story$ performance and $movie \rightarrow movie$ performance using a blockwise permutation test.

In each trial, we randomly resampled (with replacement) 10-TR blocks from the voxel's actual response time course, before taking the linear correlation between the predicted response time course and the permuted response time course. Resampling contiguous blocks preserves the auto-correlation structure of the voxel's responses. Repeating this process for 10,000 trials provided a null distribution of within-modality scores for each voxel. We identified voxels with within-modality scores that were significantly higher than this null distribution than expected by chance (q(FDR) < 0.05).

\subsection{PC correlation significance test}

We projected the language and the vision encoding model weights on each principal component (PC) of the language encoding model weights. We assessed the degree to which each PC is shared between language and vision by spatially correlating the projections of the language and the vision encoding model weights. We identified PCs with significant spatial correlations using a blockwise permutation test.

In each trial, we randomly resampled (with replacement) 10-TR blocks from the movie-fMRI responses before estimating a null vision encoding model. The projections of the language encoding model weights were correlated with the projections of the null vision encoding model weights. Repeating this process for 1,000 trials provided a null distribution of spatial correlations for each PC. We identified PCs with spatial correlations that were significantly higher than this null distribution than expected by chance (q(FDR) < 0.05).

\section{Principal components}
\label{supp:pc}

We projected stimulus features onto language PCs 2, 3, 4, and 5 to interpret the semantic dimension that each PC captures, and we projected encoding model weights onto language PCs 2, 3, 4, and 5 to assess how the corresponding semantic dimensions are represented across cortex.

\begin{figure}
  \centering
  \includegraphics[width=\textwidth]{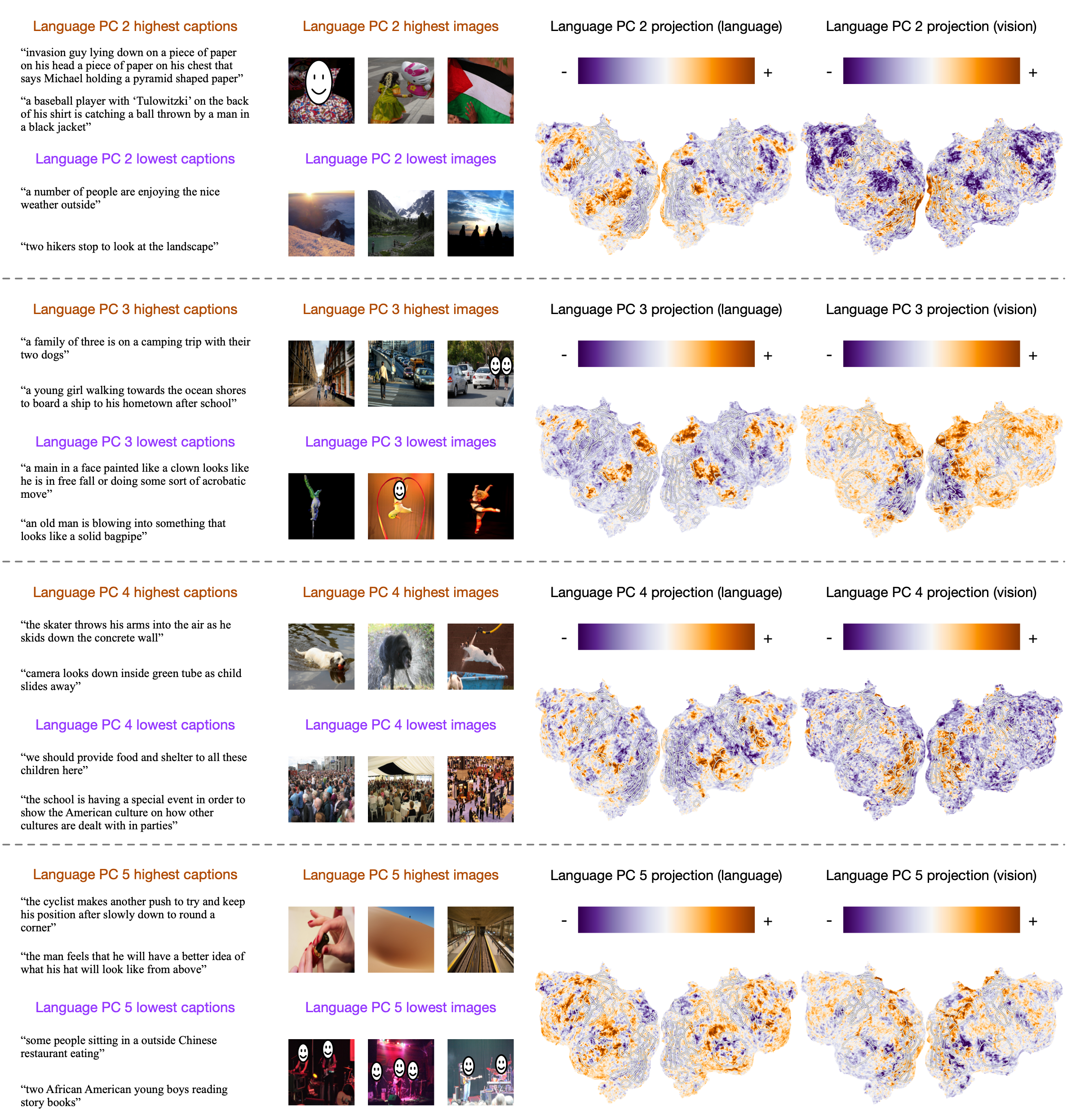}
  \caption{Interpreting language PCs 2, 3, 4, and 5.}
\label{fig:SB_pca}
\end{figure}

Figure \ref{fig:SB_pca} corresponds to Figure \ref{fig:pca} in the main text.

\newpage

\section{Flatmaps}
\label{supp:flatmaps}

We created flatmaps for subjects S2, S3, S4, S5. We visualized results by projecting scores for each voxel in a subject on that subject's cortical surface. We found similar results across subjects.

\begin{figure}[hbt!]
  \centering
  \includegraphics[width=\textwidth]{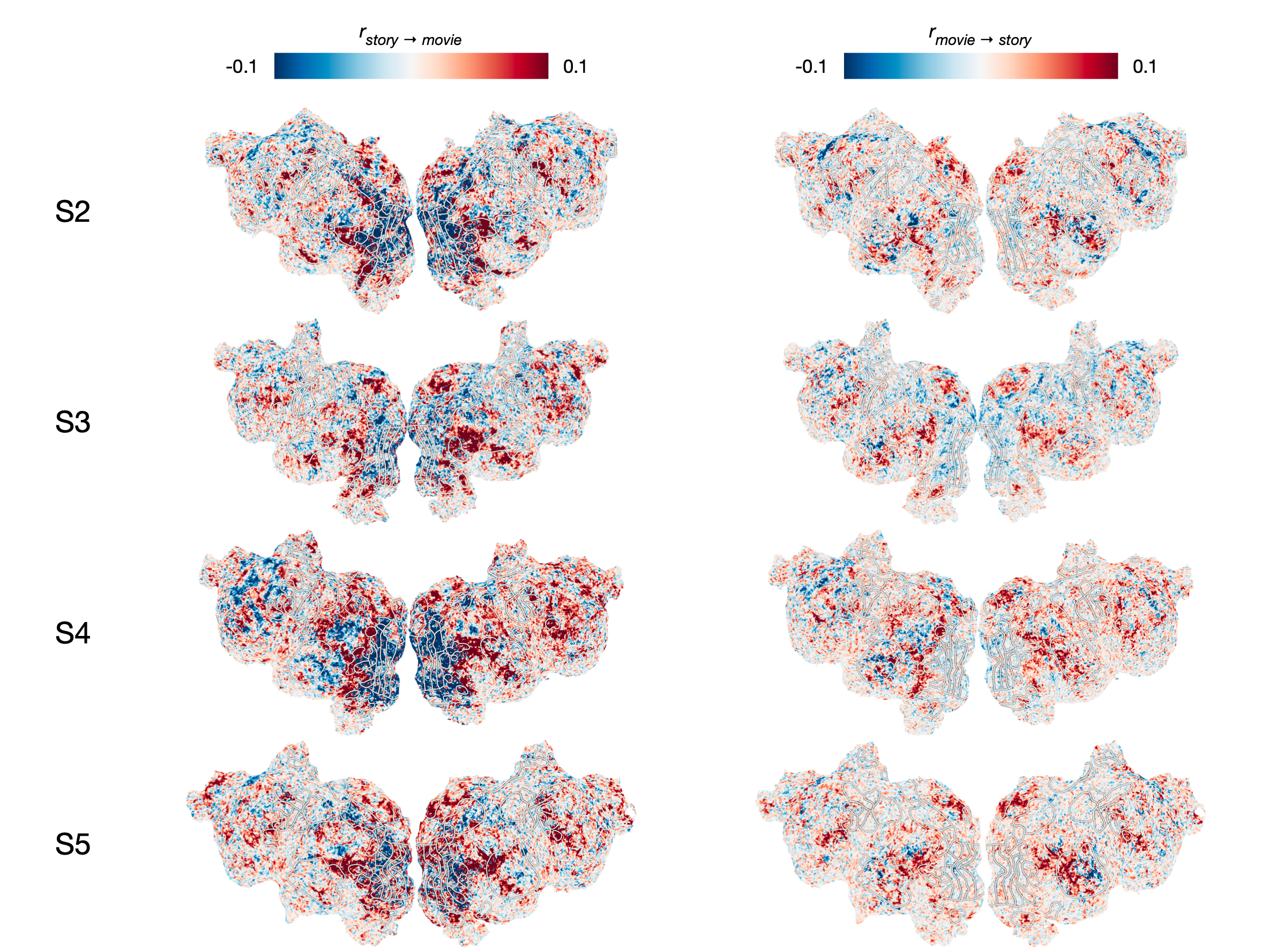}
  \caption{Cross-modality prediction performance for subjects S2, S3, S4, and S5.}
\label{fig:SC_crossmodality}
\end{figure}

\begin{figure}[hbt!]
  \centering
  \includegraphics[width=\textwidth]{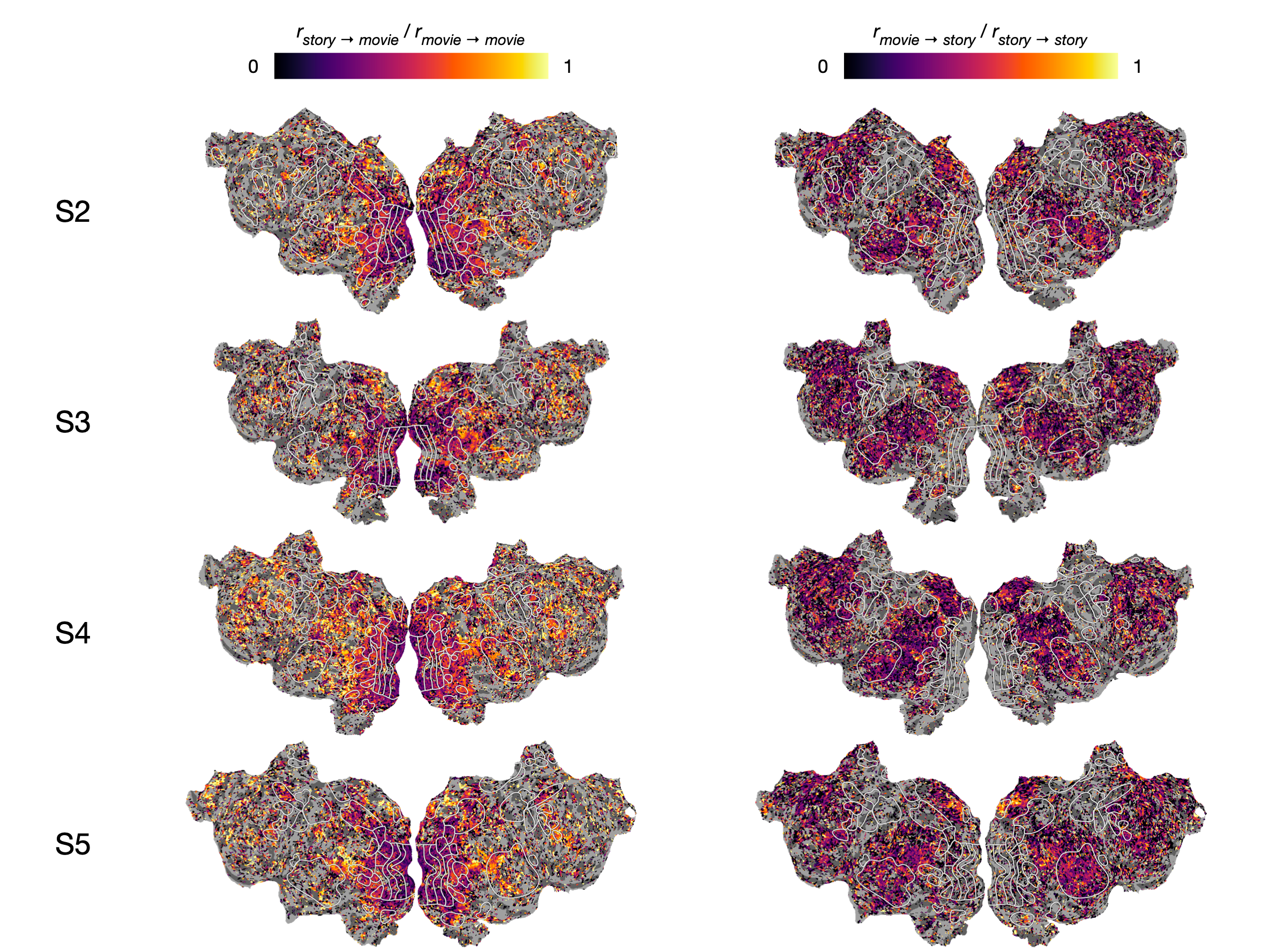}
  \caption{Comparing cross- and within-modality prediction performance for subjects S2, S3, S4, and S5.}
\label{fig:SC_withinmodality}
\end{figure}

\begin{figure}[hbt!]
  \centering
  \includegraphics[width=\textwidth]{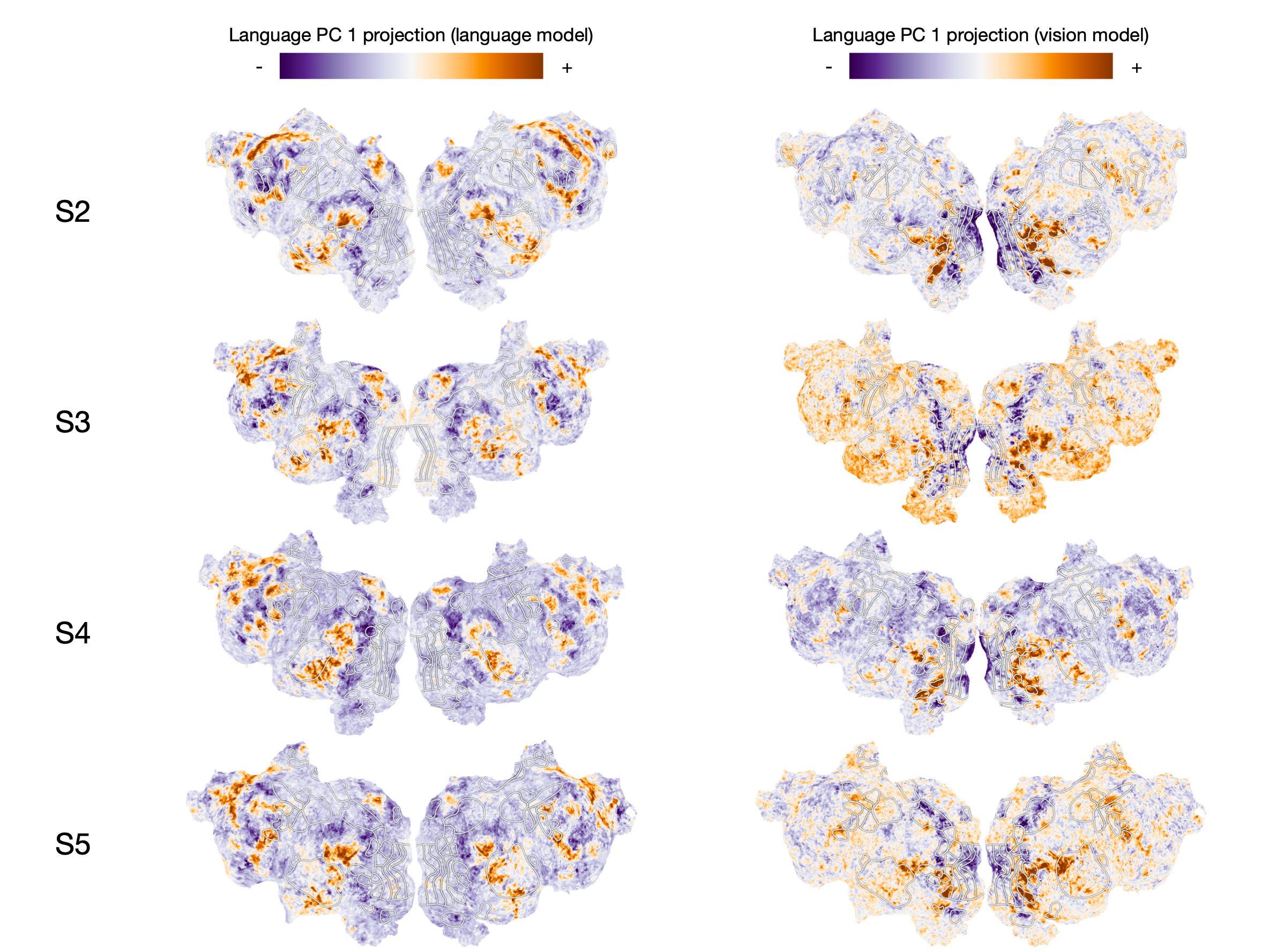}
  \caption{Language PC 1 projections for subjects S2, S3, S4, and S5.}
\label{fig:SC_pca}
\end{figure}

\begin{figure}
  \centering
  \includegraphics[width=\textwidth]{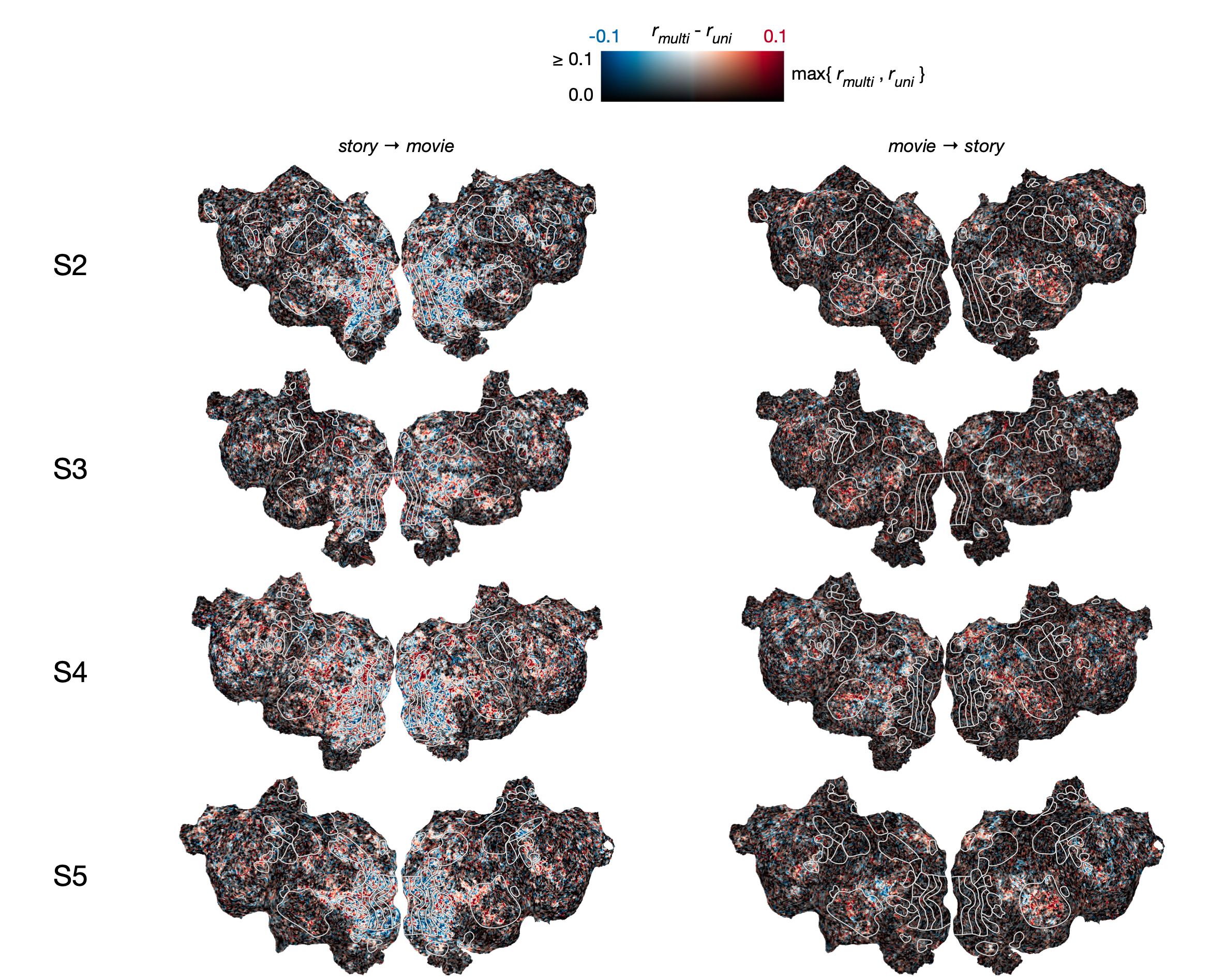}
  \caption{Multimodal and unimodal transfer performance for subjects S2, S3, S4, and S5.}
\label{fig:SC_multivsuni}
\end{figure}

Figure \ref{fig:SC_crossmodality} corresponds to Figure \ref{fig:crossmodality} in the main text. Figure \ref{fig:SC_withinmodality} corresponds to Figure \ref{fig:withinmodality} in the main text. Figure \ref{fig:SC_pca} corresponds to Figure \ref{fig:pca} in the main text. Figure \ref{fig:SC_multivsuni} corresponds to Figure \ref{fig:multivsuni} in the main text.

\newpage

\newpage

\newpage

\newpage

\section{BridgeTower-Large}
\label{supp:btlarge}

We repeated the analyses in the main text using BridgeTower-Large to extract stimulus features instead of BridgeTower-Base \cite{xu2022bridge}. For unimodal baselines, we used the pretrained RoBERTa-Large \cite{Liu2019RoBERTaAR} and ViT-Large \cite{Dosovitskiy2020AnII} transformers used to initialize BridgeTower-Large. We found similar results for BridgeTower-Base and BridgeTower Large.

\begin{figure}
  \centering
  \includegraphics[width=\textwidth]{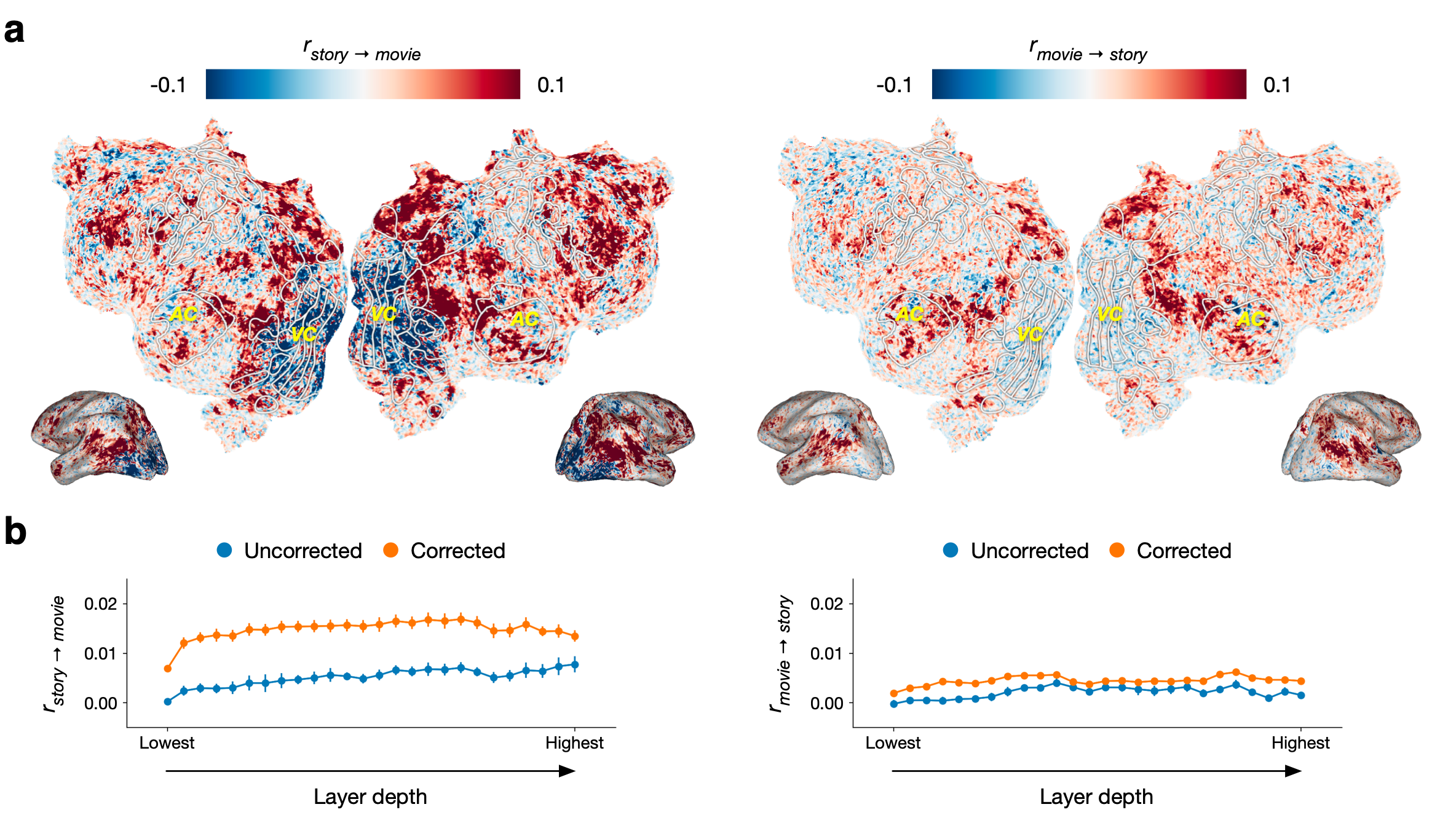}
  \caption{Cross-modality prediction performance for BridgeTower-Large.}
\label{fig:SD_crossmodality}
\end{figure}

\begin{figure}
  \centering
  \includegraphics[width=\textwidth]{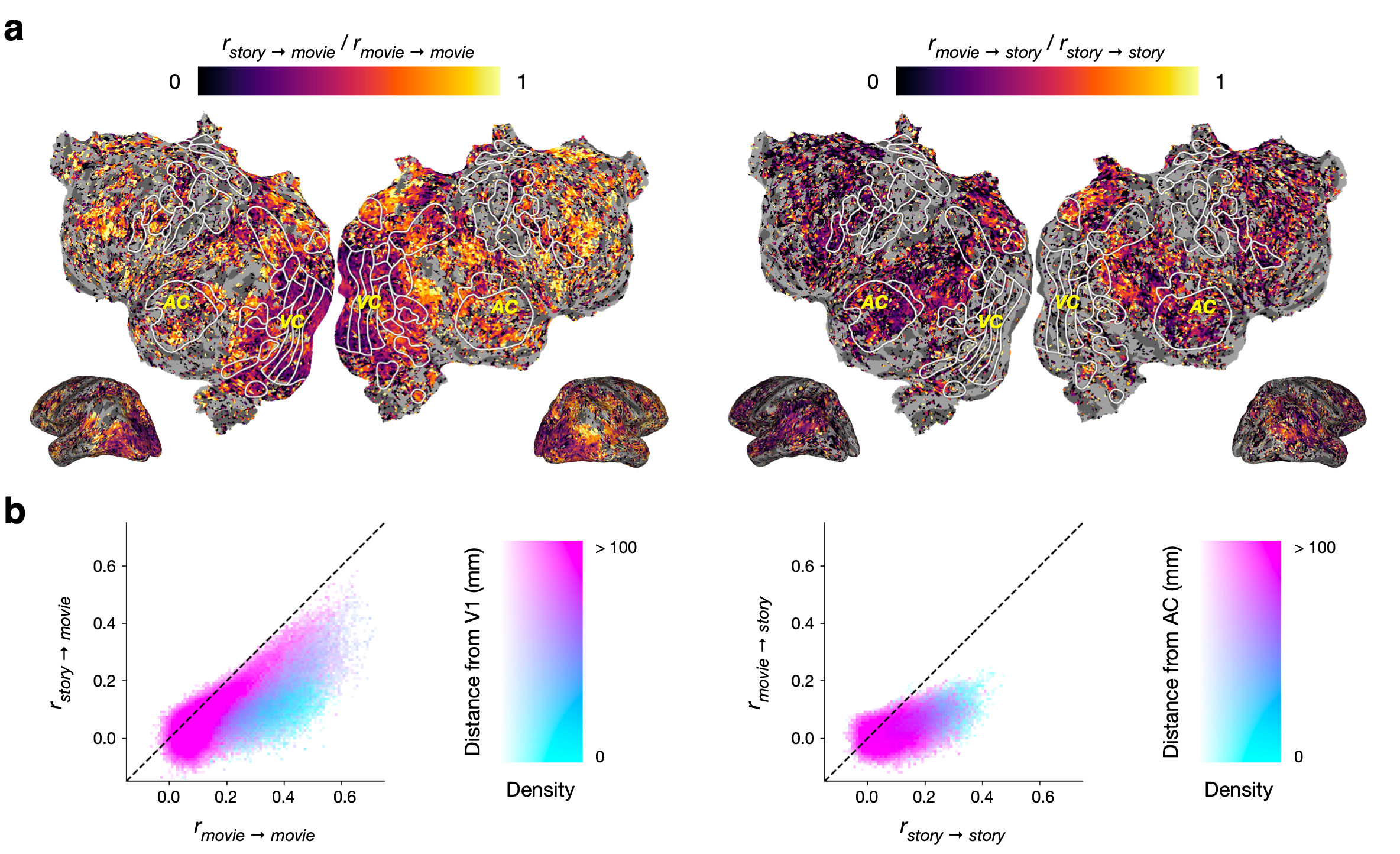}
  \caption{Comparing cross- and within-modality for BridgeTower-Large.}
\label{fig:SD_withinmodality}
\end{figure}

\begin{figure}
  \centering
  \includegraphics[width=\textwidth]{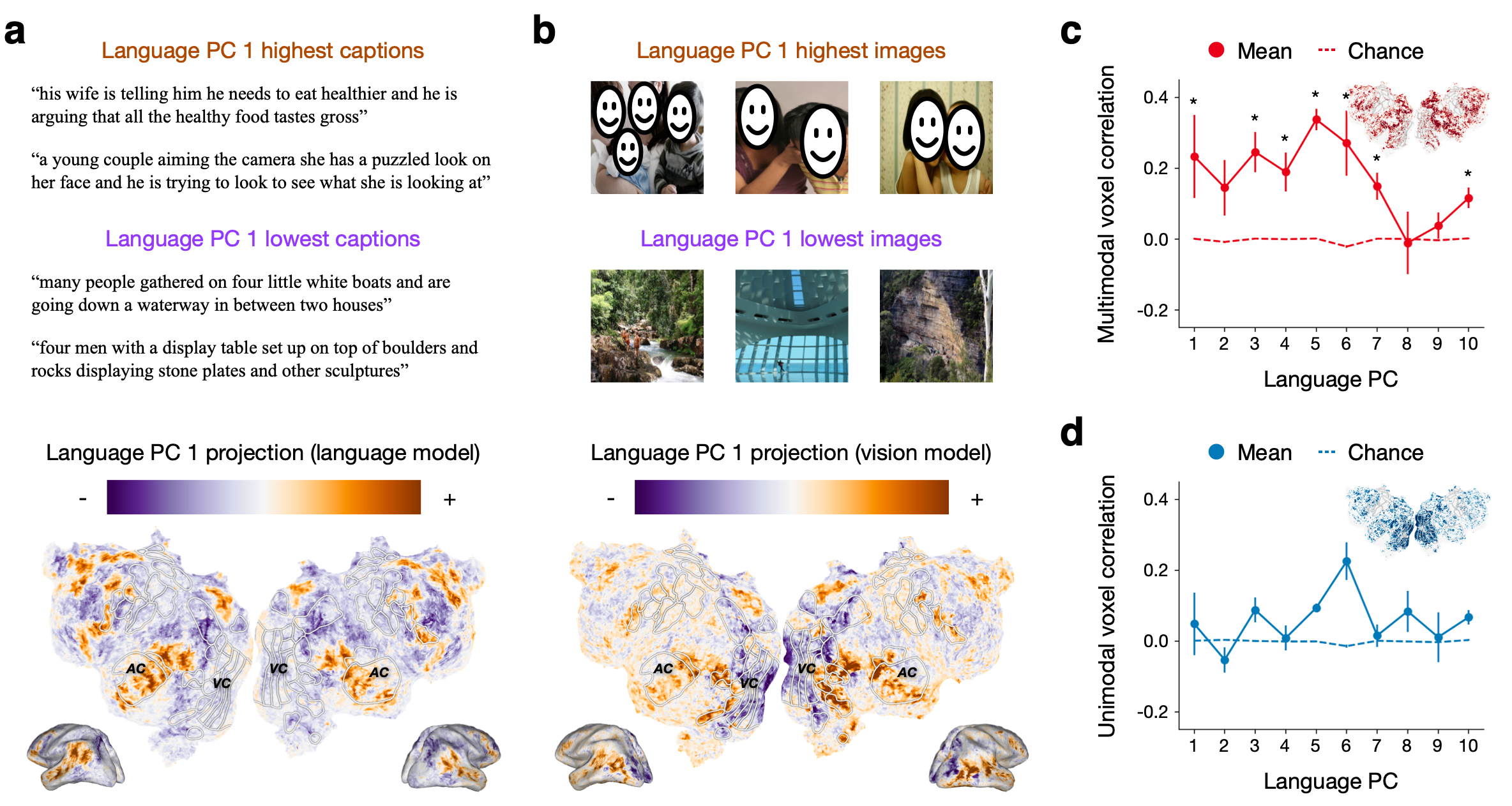}
  \caption{Encoding model principal components for BridgeTower-Large.}
\label{fig:SD_pca}
\end{figure}

\begin{figure}
  \centering
  \includegraphics[width=\textwidth]{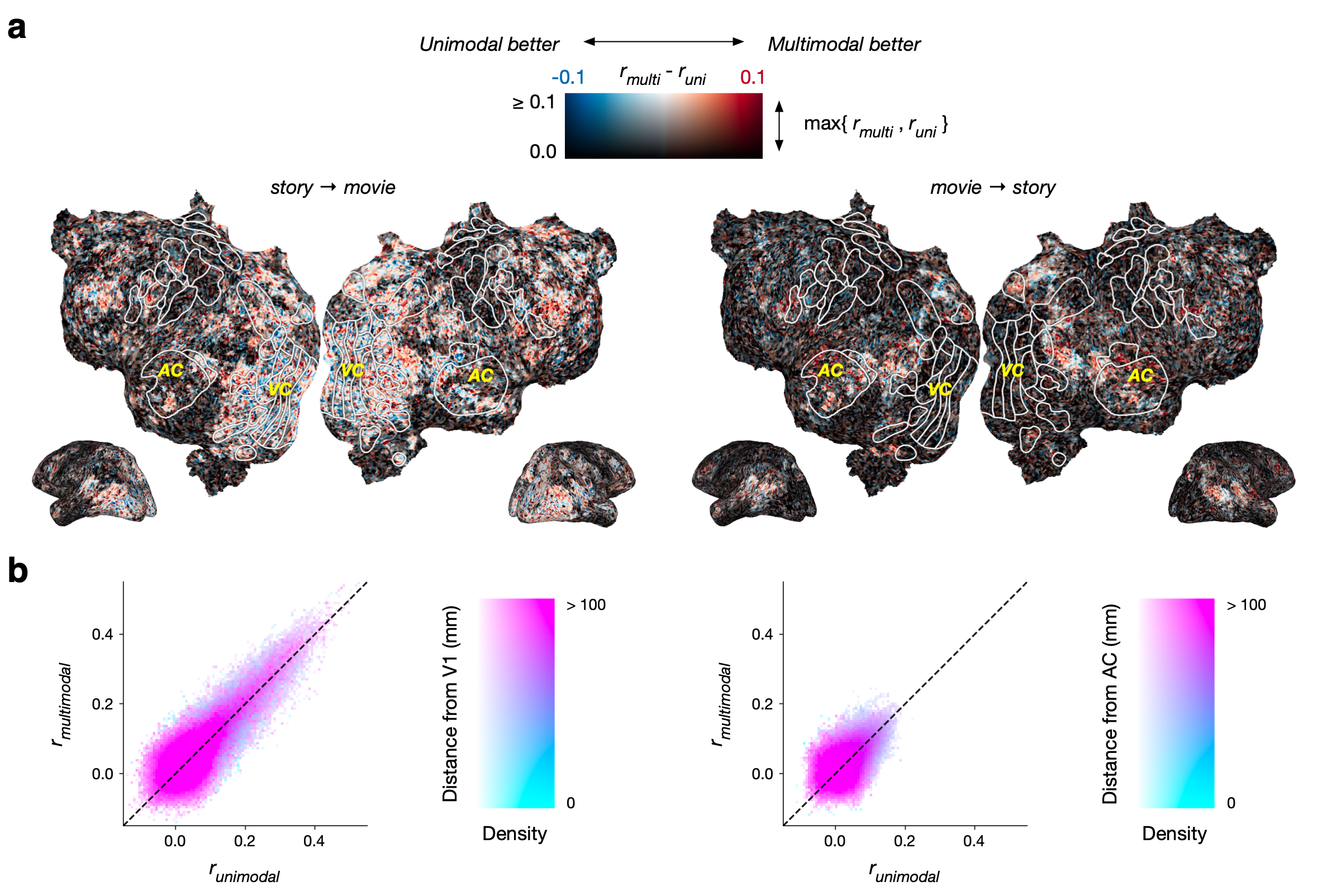}
  \caption{Multimodal and unimodal transfer performance for BridgeTower-Large.}
\label{fig:SD_multivsuni}
\end{figure}

Figure \ref{fig:SD_crossmodality} corresponds to Figure \ref{fig:crossmodality} in the main text. Figure \ref{fig:SD_withinmodality} corresponds to Figure \ref{fig:withinmodality} in the main text. Figure \ref{fig:SD_pca} corresponds to Figure \ref{fig:pca} in the main text. Figure \ref{fig:SD_multivsuni} corresponds to Figure \ref{fig:multivsuni} in the main text.

\end{document}